\documentclass[a4paper]{article}

\usepackage[english]{babel}
\usepackage[utf8x]{inputenc}
\usepackage[T1]{fontenc}

\usepackage[a4paper,top=3cm,bottom=2cm,left=3cm,right=3cm,marginparwidth=1.75cm]{geometry}
\usepackage{authblk}

\usepackage{amsmath}
\usepackage{bbm}
\usepackage{graphicx}
\usepackage{wrapfig}
\usepackage{subcaption}
\usepackage[colorinlistoftodos]{todonotes}
\usepackage[colorlinks=true, allcolors=blue]{hyperref}
\usepackage{apacite}
\usepackage{fancyhdr}
\fancypagestyle{firstpage}{
    \fancyhf{} 
    \fancyhead[C]{\fontsize{9}{12}\selectfont The 5\textsuperscript{th} Symposium on Management of Future  Motorway and Urban Traffic Systems (MFTS 2024)} 
}

\begin{document}


\title{A Sequential Decision-Making Model for Perimeter Identification}

\author{Ayal Taitler}

\affil{Department of Industrial Engineering and Management \\Ben-Gurion University Of The Negev, Israel}

\date{\vspace{-5ex}}

\maketitle
\thispagestyle{firstpage}

\renewcommand{\abstractname}{SHORT SUMMARY}

\begin{abstract}
Perimeter identification involves ascertaining the boundaries of a designated area or zone, requiring traffic flow monitoring, control, or optimization. Various methodologies and technologies exist for accurately defining these perimeters; however, they often necessitate specialized equipment, precise mapping, or comprehensive data for effective problem delineation. In this study, we propose a sequential decision-making framework for perimeter search, designed to operate efficiently in real-time and require only publicly accessible information. We conceptualize the perimeter search as a game between a playing agent and an artificial environment, where the agent's objective is to identify the optimal perimeter by sequentially improving the current perimeter. We detail the model for the game and discuss its adaptability in determining the definition of an optimal perimeter. Ultimately, we showcase the model's efficacy through a real-world scenario, highlighting the identification of corresponding optimal perimeters.
\end{abstract}


\section{INTRODUCTION}
In urban transportation systems, effective traffic management is paramount to ensuring safety, reducing congestion, and enhancing overall mobility. A critical aspect of traffic engineering involves accurately identifying perimeters within which traffic conditions are to be monitored, controlled, or optimized. The challenge of perimeter identification in traffic engineering revolves around delineating the boundaries of specific areas or regions where traffic characteristics are of particular interest. These areas may span from city centers and neighborhoods to highway sections and intersections. Precise and fast perimeter identification is crucial for deploying targeted traffic management solutions tailored to the unique needs and challenges of each defined zone.


There are primarily two categorizations for perimeter identification: static and dynamic. Static identification usually focuses on analyzing the connectivity and parameters of the network, forming partitions typically based on optimization or clustering techniques. For instance, \citeA{jiang2023partitioning} (2023) employed a Graph Convolutional Network (GCN) \cite{zhang2019graph} to integrate spatial connectivity and traffic variables, subsequently identifying three levels of clusters. \citeA{liu2019new} introduced a weighted average between flow and speed, constructing a MILP (Mixed-Integer Linear Programming) model to identify the major skeleton of the perimeter.
Dynamic identification involves addressing changing traffic measures over time. A popular approach is to update the historical perimeter boundaries based on new incoming data. For example, \citeA{ji2014empirical} applied boundary adjustment using a dynamic mechanism for the city of Shenzhen, China. \citeA{saeedmanesh2017dynamic} identified clusters via the snake clustering algorithm \citeA{saeedmanesh2016clustering} and skeleton via MILP, subsequently refining them through another MILP and dynamic clustering.

In this work, we formulate the perimeter identification problem as a sequential decision-making model based on congestion heat map images that can be obtained in real-time from sources such as Google Maps (GMaps) \cite{gmaps}. Our model decomposes the global optimization inherent in the perimeter identification problem into discrete sequential steps, each aimed at guiding a "playing" agent toward the optimal solution. This approach enables the solving agent to comprehend the problem's structure and facilitates the transfer of understanding between identification problems. Additionally, it allows for tracking dynamic changes in the perimeter as part of its gameplay.

\section{METHODOLOGY}
The Markov Decision Process (MDP) \cite{puterman1990markov} is a widely adopted approach for modeling sequential decision-making processes. Here, we model the perimeter identification problem as a game between two players. The first player is the environment, representing the current state of congestion, the existing perimeters, and how they can be altered. The second player is the agent, whose goal is to determine the best possible perimeter. The agent executes a sequence of actions, with each action altering the state of the environment, i.e., the active perimeter.

\subsection{MDPs Background}
An finite-horizon MDP is a tuple $\langle S,A,R,T\rangle$, where $S$ is the state space; $A$ is the action space; $T(s'|s,a)$ is the probability that the system transitions to state $s'$ given state $s$ and action $a$; and $R(s,s')$ is the immediate reward function. The fact that $T(s'|s,a)$ is a function of $s,\ s'$ and $a$ only (and not the previous trajectory of the system) is referred to as the Markov property. We denote the horizon of the process as $H$.

A policy $\pi:S \rightarrow \mathcal{P}(A)$ is a mapping from the state space to the set of probability distributions over the action space. The probability of taking action $a$ given state $s$ is denoted by $\pi(s|a)$. The optimal policy is denoted by $\pi^*(s|a)$. The state–action value function $Q^\pi(s,a)$ of a policy $\pi$, namely the expected total reward starting from a state $s$, taking action $a$, and following $\pi$ afterward, is defined as the value of 
$\pi$:
\begin{equation} \label{eq:Q_def}
    Q^\pi(s,a) = E_{\pi} \bigg[ \sum_{k=0}^{H}{r_{t+k+1} | s_t=s, a_t=a} \bigg] = R(s,s') + \sum_{s\in S}T(s'|s,a)\sum_{a'\in A}\pi(a'|s')Q^{\pi}(s',a')
\end{equation}
The optimal state-action-value function $Q^*(s,a)$ of $\pi^*$, namely the expected total reward starting from a state $s$, taking action $a$, and acting optimally afterwards is defined as
\begin{equation} \label{eq:Q_star}
    Q^*(s,a) = R(s,s') + \max_{a'}\sum_{s\in S}T(s'|s,a)Q^*(s',a')
\end{equation}

Full evaluation of the optimal state-action value function $Q^*$ requires intractable computation over the whole action and state spaces. Thus, an approximate version to estimate $Q$, known as \textit{Q-learning} \cite{sutton2018reinforcement}, try to find the optimal policy by fitting the Q-function with temporal difference (TD) learning (that during each training iteration perturbs the Q-function toward the one-step bootstrapped estimation that is identified by the Bellman function):
\begin{equation} \label{eq:q_learning}
    Q'(s,a) \leftarrow (1-\alpha)Q(s,a) + \alpha \Big(R(s,s') + \max_{a'} Q(s',a')\Big)
\end{equation}
where $\alpha$ is referred to as the learning rate.

\subsection{The Markov Model} \label{sec:mdp}
To define the perimeter identification model as a sequential model, we will frame it as a search problem. The objective of the search is to locate the intersections, acting as the vertices of the perimeter. Here, we make the assumption that the perimeter shape is convex. This assumption is not overly restrictive, as the search process can be carried out iteratively to identify the perimeter as a combination of convex shapes. The MDP that defines the search problem is as follows:

\paragraph{States $S$} The state of the system comprises the current selection of intersections, which determines a convex hull defining the perimeter. Clearly, intersections within the set residing inside the convex hull do not influence the shape of the perimeter but rather define a distinct state within the state space.

\paragraph{Actions $A$} The actions in the game are straightforward: either add a new intersection to the set of intersections or remove an existing intersection from the current set. Once more, adding a new intersection that lies within the induced convex hull of the current state will transition the problem to a new state, yet it will not alter the actual convex hull defining the perimeter.

\paragraph{Transition $T(s'|s,a)$} The transition function represents the "dynamics" of the game, defining how the system's state will transition from a particular state given an action. Therefore, in this scenario, the transition function merely modifies the set of currently selected intersections. The convex hull induced by the set of intersections is a direct representation of it, hence it is unnecessary for defining the state. However, it may prove useful for a more detailed state representation, as will be discussed in Section \ref{sec:representation}.

\paragraph{Reward $R(s,s')$} The reward represents the effectiveness of the transition, indicating how much closer the optimization has come to achieving the optimal perimeter. Therefore, the reward signifies the additional portion of the perimeter beyond the previous perimeter. Specifically, the reward corresponds to the area added to the perimeter, which is proportional to the congestion in that area. To prevent the optimization from simply covering the entire field of view, penalties for adding non-congested areas are subtracted as a regularization measure for the optimization. The explicit reward expression is given in equation \eqref{eq:reward}.
\begin{equation} \label{eq:reward}
    R(s,s') = V(s')-V(s)= \frac{1}{\beta}\bigg(\sum_{p \in CH(s')}w_p - \lambda\mathbbm{1}_{\{w_p=0\}} - \sum_{p \in CH(s)}w_p - \lambda\mathbbm{1}_{\{w_p=0\}}\bigg)
\end{equation}
where $\beta$ is a normalization term to scale the reward, e.g., the area of the frame. $\lambda$ is the designer chosen regularization term for non-congested areas, $CH(s)$ is the convex hull of the heat map of state $s$, and $p$ represents a pixel belonging to the convex hull.
Lastly, $w_p$ is the weight or level of congestion, of pixel $p$. 

\subsection{States and Actions Representations} \label{sec:representation}
State representation is a critical aspect of this model. In a simple representation of the state as an ordered set of intersections, the playing agent attempts to optimize the perimeter based solely on the reward. That is, all information about the game -- such as the appearance of congested areas and the topological changes resulting from adding intersection $i$ to the state -- is inferred through the reward feedback. Consequently, altering the target area or even the distribution of congestion yields a different reward signal with the same state and action spaces, requiring a learning agent to essentially start from scratch in finding the optimal perimeter. A more elaborate state representation, capable of capturing topological properties, can be beneficial for generalization or knowledge transfer between different levels of congestion or even cities, facilitating the automatic discovery of perimeters.

The actions in the game play a significant role, both in terms of computational complexity and the quality of the perimeter. Too many actions may lead to an intractable optimization problem, while too few may limit the representation quality of the perimeter by the selected intersections. Additionally, the selected actions should undergo a connectivity analysis to ensure that the identified perimeter is a connected component in the city network. In the context of this work, we only demonstrate the optimization game using a basic set of actions to illustrate the technical merits of the model.

\section{RESULTS AND DISCUSSION}

\begin{wrapfigure}{R}{0.45\textwidth}
\centering
\includegraphics[width=0.39\textwidth]{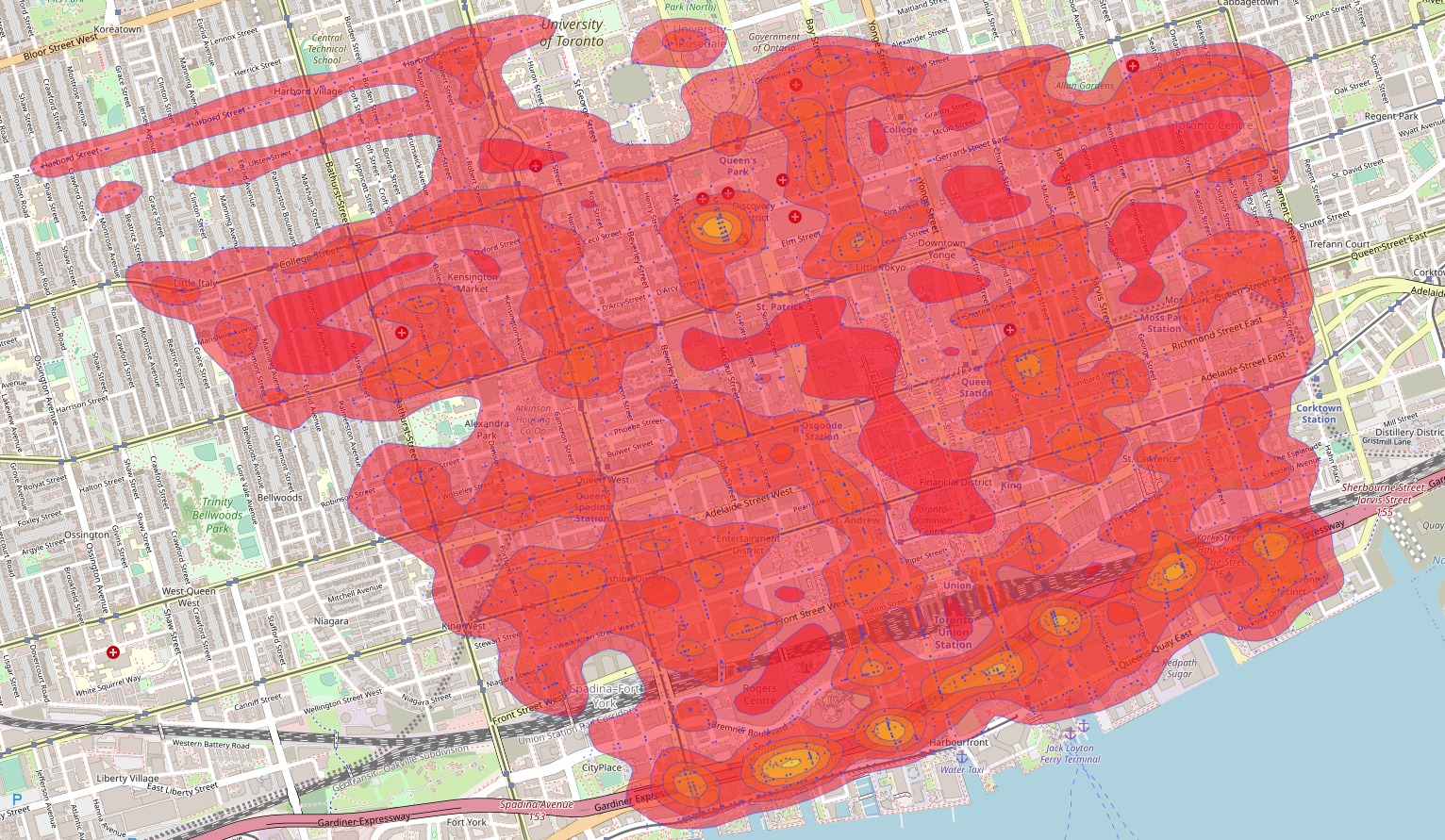}
\caption{Toronto downtown heat-map.}
\label{fig:toronto}
\end{wrapfigure}
We used here Google Maps \cite{gmaps} as it gives congestion heat maps over a chosen area map.
The heat map layer is updated online based on traffic data collected from sensors and drivers in real time. Thus, it is useful both for static and dynamic perimeter identification. Here we focus on a identifying a single perimeter, and show how it can be easily identified for a given heat map using the MDP presented in section \ref{sec:mdp}. For the playing agent we used a Q-Learning \cite{sutton2018reinforcement} agent implementing the equation in \eqref{eq:q_learning}. We look at downtown Toronto, which is a highly congested area. The heat map of downtown Toronto is given in figure \ref{fig:toronto}.

Given the selected area's image, the heat map layer and the intersections can be identified using classical image processing. The information can also be extracted using other tools to get an accurate network of intersections, which serves as the actions in the game. Here, to illustrate the effectiveness of the game and optimization we extracted everything from the image directly.

\begin{figure}
\centering
\begin{subfigure}[b]{0.32\textwidth}
   \includegraphics[width=1\linewidth]{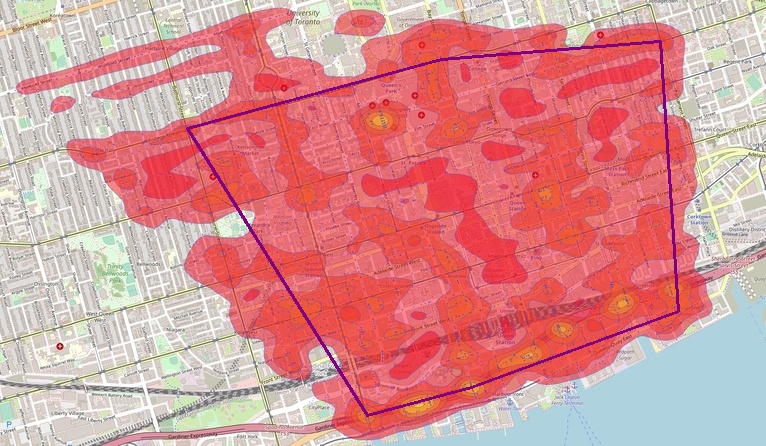}
   \caption{Conservative perimeter identification with $\lambda=10$.}
   \label{fig:Ng1} 
\end{subfigure}
\begin{subfigure}[b]{0.32\textwidth}
   \includegraphics[width=1\linewidth]{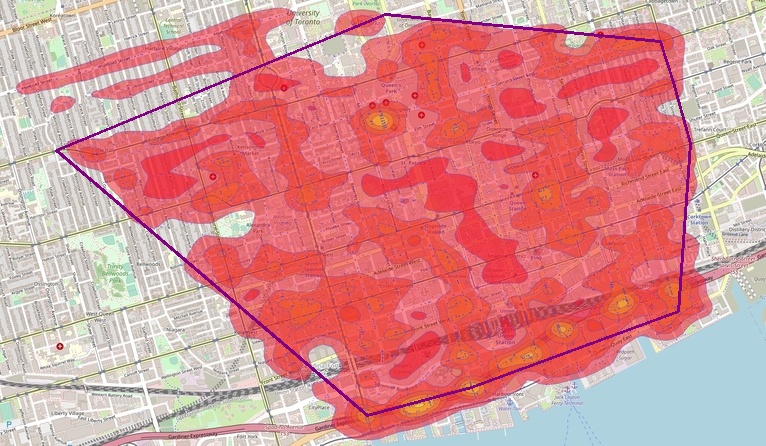}
   \caption{Balanced perimeter identification with $\lambda=1$.}
   \label{fig:Ng2}
\end{subfigure}
\begin{subfigure}[b]{0.32\textwidth}
   \includegraphics[width=1\linewidth]{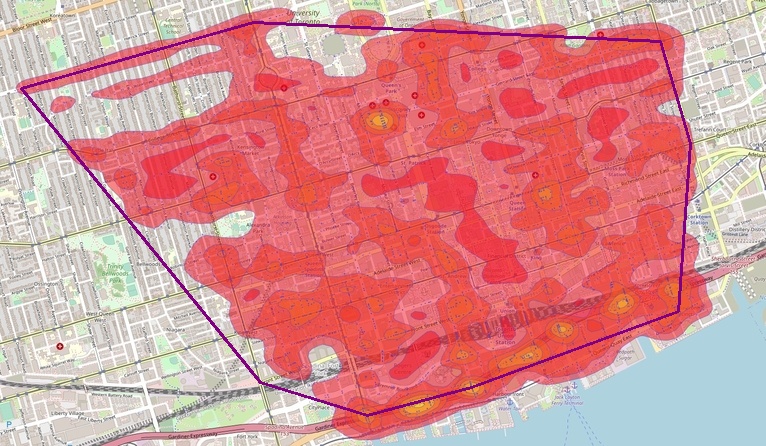}
   \caption{Non-conservative perimeter identification with $\lambda=0.1$.}
   \label{fig:Ng2}
\end{subfigure}
\caption{Identified perimeters from three different games.}
\label{fig:results}
\end{figure}

We conducted three games for our Q-Learning agent.
\begin{itemize}
    \item Conservative game: The objective of this game was to identify the smallest perimeter encompassing solely the congested areas, excluding congested forks with uncongested areas. Therefore, in this game, the agent received significant penalties for including uncongested areas. In this case, the regularization term was set to $\lambda=10$.
    \item Balanced game: The objective here was to strike a balance between incorporating uncongested areas and including congested areas. In essence, the agent could include uncongested areas if doing so enabled it to encompass more of the congestion within the perimeter. The regularization term was set to $\lambda=1$ in this case.
    \item Non-conservative game: The aim here was to incentivize the agent to incorporate all congestion, even if it entailed including non-congested regions. In this case, the regularization term was set to $\lambda=0.1$.
\end{itemize}
In all three games, the initial state was the same random state, chosen to demonstrate the varying evolution based on the optimization requirements. The results for the three games are presented in Figure \ref{fig:results}. It can be observed that the agent was able to quickly identify the appropriate perimeters for different regularization terms. This highlights the flexibility of the model in easily adapting to various needs.

At this point, it's important to note that the convex hulls found are not the actual perimeters; rather, they are convex hulls with intersections as their vertices. To generate an actual perimeter, a deterministic post-processing step is necessary, based on the network's connectivity and edges of the convex hull.

\section{CONCLUSIONS}
We have introduced a sequential decision-making model for identifying perimeters in congested regions. This model frames the identification problem as a game, where a player selects the vertices of a convex hull to define the perimeter. Essentially, the proposed game solves the optimization task of finding the optimal perimeter. Additionally, we have introduced a regularization term, serving as a control mechanism for the model designer to adjust and fine-tune perimeter requirements directly into the optimization process.

By formulating the optimization as a sequential model, we enable the utilization of learning tools. This approach facilitates knowledge transfer and generalization of perimeters across different cities. Furthermore, it allows for tracking a dynamic perimeter without the need to solve new optimization problems from scratch.

\newpage
\bibliography{sample}
 
\end{document}